\documentclass[letterpaper, 10 pt, conference]{ieeeconf}  

\IEEEoverridecommandlockouts                              

\usepackage{graphics} 
\usepackage{epsfig} 
\usepackage{times} 
\usepackage{amsmath} 
\usepackage{amssymb}  
\usepackage{hyperref}
\usepackage{booktabs}
\newcommand{\thickhline}{\specialrule{0.8pt}{0pt}{0pt}}
\usepackage{siunitx}
\usepackage[utf8]{inputenc}
\usepackage{tikz}
\usepackage{tikz-3dplot}

\usepackage{pgfplots}

\usepackage[left=0.67in, right=0.67in, top=0.8in, bottom=0.65in]{geometry}

\DeclareUnicodeCharacter{2212}{−}
\usepgfplotslibrary{groupplots,dateplot}
\usetikzlibrary{patterns,shapes.arrows}
\pgfplotsset{compat=newest}

\usepackage{pgf}
\usetikzlibrary{arrows}
\usetikzlibrary{arrows.meta} 
\usetikzlibrary{shapes.multipart,positioning}
\usepackage{pgfplots}
\pgfplotsset{compat=1.17}

\newcommand{\robotname}{Jumper}

\title{\LARGE \bf
Towards Quadrupedal Jumping and Walking for Dynamic Locomotion using Reinforcement Learning
}

\author{Jørgen Anker Olsen,  Lars Rønhaug Pettersen, and Kostas Alexis
\thanks{The authors are with the Autonomous Robots Lab, NTNU, O.S. Bragstads Plass 2D, 7034, Trondheim, Norway. {\tt\small jorgen.a.olsen@ntnu.no}}%
}

\begin{document}

\maketitle
\thispagestyle{empty}
\pagestyle{empty}

\begin{abstract} This paper presents a curriculum-based reinforcement learning framework for training precise and high-performance jumping policies for the robot `\robotname{}'. Separate policies are developed for vertical and horizontal jumps, leveraging a simple yet effective strategy. First, we densify the inherently sparse jumping reward using the laws of projectile motion. Next, a reference state initialization scheme is employed to accelerate the exploration of dynamic jumping behaviors without reliance on reference trajectories. We also present a walking policy that, when combined with the jumping policies, unlocks versatile and dynamic locomotion capabilities. Comprehensive testing validates walking on varied terrain surfaces and jumping performance that exceeds previous works, effectively crossing the Sim2Real gap. Experimental validation demonstrates horizontal jumps up to \SI{1.25}{\meter} with centimeter accuracy and vertical jumps up to \SI{1.0}{\meter}. Additionally, we show that with only minor modifications, the proposed method can be used to learn omnidirectional jumping.

\end{abstract}

\section{INTRODUCTION}

Quadruped robots can navigate complex terrains and overcome obstacles not only through walking but also through powerful jumps. The combination of robust walking and precise jumping capabilities is particularly valuable for planetary exploration~\cite{arm2023scientific,spiridonov2024spacehopper}. In reduced gravity environments, a quadrupedal robot could use walking for normal traversal and jumping to overcome obstacles much larger than itself~\cite{olsen2025olympus}. 

Deep Reinforcement Learning (DRL) has proven to be a powerful tool for training quadrupedal walking policies that demonstrate unprecedented robustness and capabilities compared to non-learning-based methods \cite{rudin2025parkour}. Jumping presents a more challenging dynamic task requiring precise coordination, while operating at the physical limits of the robot hardware. Additionally, due to extended periods of low control authority (e.g., during the ``flight'' phase), jumping controllers are required to plan ahead over a substantial horizon. This has traditionally motivated the use of trajectory optimization for controller synthesis~\cite{nguyen2019optimized}. The corresponding optimization problems suffer from high dimensionality and non-linearity, making them challenging to solve and numerically intensive~\cite{nguyen2022contact}. Multiple simplifications are often necessary to make the problem tractable, which compromises performance. 

Learning-based approaches overcome the need for model simplifications by instead learning directly from environment interaction. However, when encountering complex dynamics and sparse rewards, learning algorithms struggle to avoid local minima, where only parts of the reward are optimized. As demonstrated in~\cite{atanassov2024curriculumbased}, DRL performs poorly for quadrupedal jumping with random environment interaction alone, but succeeds when tailoring the training environment using auxiliary rewards and a curriculum. In this work, we demonstrate how to systematically tailor the training environment to obtain precise, versatile, and robust jumping policies using DRL. The method leverages projectile motion equations and enhances the exploration capabilities of the agent through an elaborate reference state initialization scheme.

We demonstrate that DRL enables quadrupeds to perform powerful and precise jumping to target heights and positions, as well as walking locomotion, through experimental validation under Earth gravity conditions. These combined capabilities position robots like \robotname{} (Fig. \ref{fig:front_figure}) for scenarios requiring versatile and dynamic locomotion. The contributions of this work include:

\begin{figure}[t]
    \centering
    \includegraphics[width=0.45\textwidth]{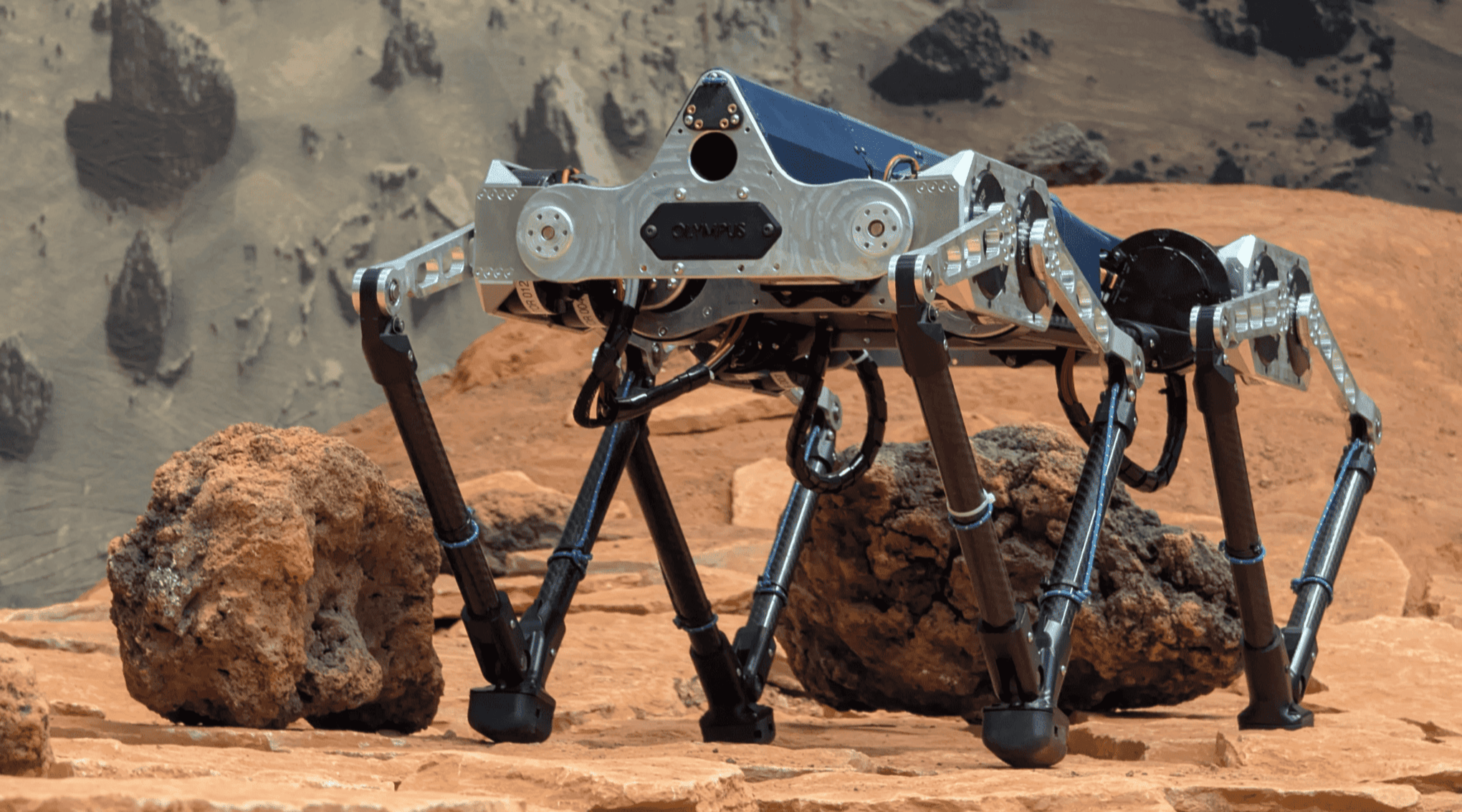}   
    \vspace{-1.5ex}    
    \caption{The \robotname{} quadruped standing in a Mars analog environment.}
    \label{fig:front_figure}
    \vspace{-4ex}
\end{figure}

\begin{itemize}
    \item A high-performing Reinforcement Learning (RL) policy for horizontal jumping that operates without reference trajectories, achieving stable landings and demonstrating jumps up to \SI{1.25}{\meter}, with centimeter accuracy.
    \item A vertical jumping policy capable of reaching specified heights with precision, tested up to \SI{1.0}{\meter} in height.
    \item A walking policy trained via reinforcement learning and validated on diverse terrain surfaces using a custom Visual-Inertial Odometry (VIO) hardware setup.   
\end{itemize}

The remainder of this paper is organized as follows: Section \ref{Sec:related_work} reviews related work. Section \ref{sec:robot_system} describes the \robotname{} quadruped. Section \ref{sec:sim_kin} details the kinematic constraints. Section \ref{sec:DRL_pipeline} outlines the training methodology. Section \ref{sec:sim_studies} shows simulation results, while Section \ref{sec:experimental_validation} presents experimental validation. Section \ref{sec:conclusion} concludes the work.

\vspace{-0.5ex}

\section{RELATED WORK}
\label{Sec:related_work}

Traditionally, legged locomotion for quadrupeds has relied heavily on model-based approaches~\cite{raibert1986legged}, including trajectory optimization~\cite{neunert2017trajectory} and model predictive control~\cite{di2018dynamic}. However, in recent years, reinforcement learning has taken the stage and proven its robustness and versatility, enabling traversal of unstructured terrain~\cite{rudin2022learning} and even parkour~\cite{rudin2025parkour}. 

The main body of the literature on jumping quadrupeds consists of model-based methods where trajectory optimization is deployed to synthesize jumping maneuvers \cite{nguyen2019optimized}. Due to high system complexity, these methods often require offline computation, specialized solvers, and advanced low-level controllers to compensate for model inaccuracies, reducing their generalizability and versatility~\cite{nguyen2022contact}. However, in conjunction with the development of learning-based methods, recent years have shown great improvement in model-based methods, enabling robust real-time controllers. This was demonstrated in \cite{Cafe_Mpc_2025} where an MIT Mini Cheetah performed a barrel roll using a model-based controller. In addition, contact implicit methods \cite{kim2025contact} represent a promising direction to remove the need for a human-provided gait schedule.

For learning-based quadruped control, two main approaches have emerged. Imitation learning leverages reference trajectories captured from real demonstrations or synthesized using classical approaches~\cite{xiao2025stable}. This makes the diversity and quality of the reference data a major concern, which naturally constrains the generalization and versatility of these methods~\cite{li2023learning}. Pure reinforcement learning approaches avoid these limitations by learning directly from environmental interaction~\cite{rudin2022learning, lee2020learning}. As these learning-based methods mature, researchers have begun to explore their potential for extreme environments, including planetary exploration \cite{arm2023scientific}, where reduced gravity amplifies jumping performance and enables traversal of obstacles larger than the robot itself, which would be impossible through walking alone~\cite{spiridonov2024spacehopper, olsen2025olympus}. Recent advances in curriculum-based RL have demonstrated complex dynamical behaviors, with~\cite{atanassov2024curriculumbased} showing the potential for pure RL-based policies capable of forward and sideways jumps, through task-based curriculum and reward shaping. Their approach employs a three-stage curriculum that requires learning vertical jumping first before progressing to horizontal jumping, using specialized rewards, including squat rewards. This was driven by the fact that direct training of horizontal jumping was found to be ineffective in learning proper jumping behaviors. In contrast, our method uses reference state initialization and the projectile motion equations to densify sparse rewards, enabling direct training of horizontal jumping without sequential curriculum behavior stages. Furthermore, we demonstrate greater jump distances and accuracy alongside the potential for omnidirectional capabilities.

\section{EXPERIMENTAL PLATFORM}
\label{sec:robot_system}

\begin{figure*}[t]
    \vspace{0ex}
    \centering    
    \includegraphics[width=0.995\textwidth]{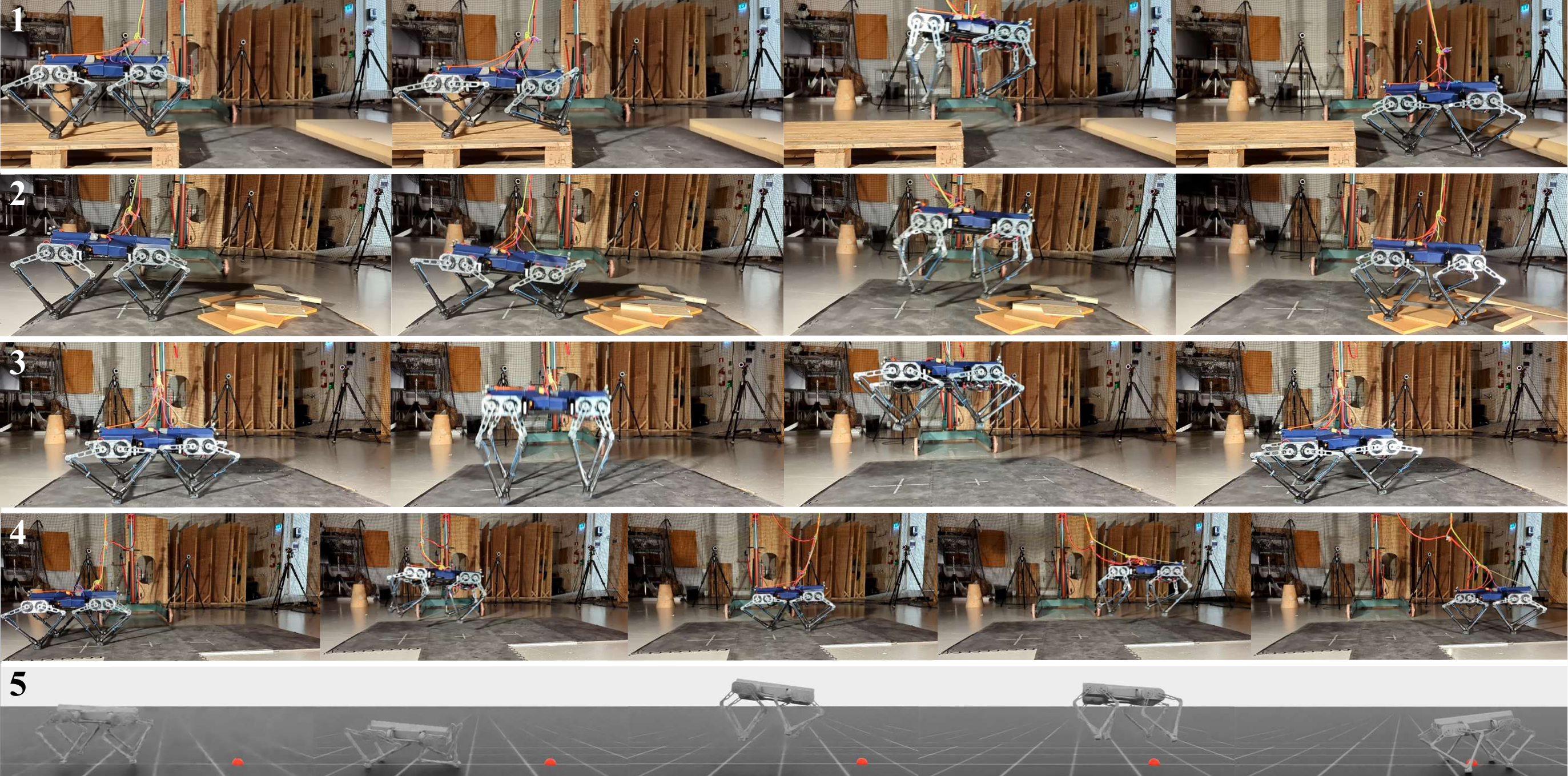}  
    \vspace{-4.5ex}
    \caption{The robot performing: 1) Jump down from  \SI{0.15}{\meter}, 2) Landing on uneven ground, 3) Vertical jump of \SI{0.75}{\meter}. 4) Sequential jumps of \SI{0.85}{\meter} forward and alternating side component of \SI{0.2}{\meter}, 5) \SI{1.25}{\meter} jump in simulation. See supplementary video for full sequences, including maximum jumps.}
    \label{fig:composit_figure}
    \vspace{-2.75ex}
\end{figure*}

For experimental validation, the \robotname{} quadruped platform was used, shown in Fig. \ref{fig:front_figure} and  Fig. \ref{fig:composit_figure}. This robot configuration was selected for its jumping capabilities, particularly its design optimized for reduced gravity locomotion using walking and jumping~\cite{olsen2025olympus}. Table \ref{tab:robot_params} lists the key robot specifications, with main characteristics including high torque and fast actuators integrated into a 5-bar linkage leg design. 

\begin{figure}[t]   
    \centering    
    \includegraphics[width=0.43\textwidth]{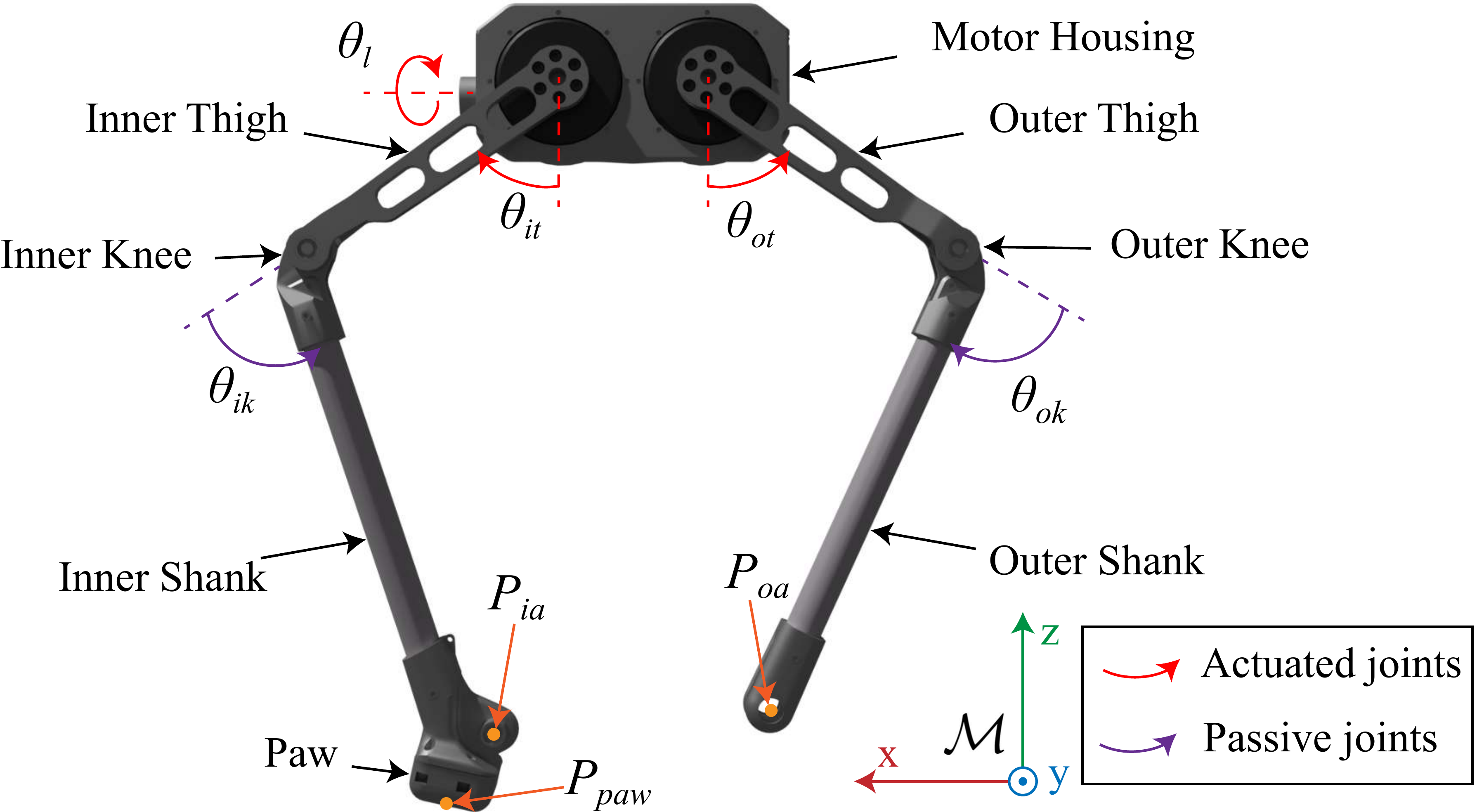}  
    \vspace{-2ex}
    \caption{Illustration of \robotname{}'s leg configuration, with key points, components, and angles. This exact leg is the left back leg.}
    \vspace{-3.75ex}
    \label{fig:leg_config}
\end{figure}

The robot's leg design, shown in Fig. \ref{fig:leg_config}, consists of three actuated and three unactuated joints per leg. Each leg's actuated joints include: the lateral motor joint $\theta_l$ that connects the motor housing to the robot base, and the inner and outer transversal motor joints $\theta_{it}$ and $\theta_{ot}$ that connect the respective thighs to the motor housing. Each leg also contains three passive joints: the inner knee $\theta_{ik}$, the outer knee $\theta_{ok}$, and the ankle joint. We define the following notation: $\boldsymbol{\theta}_m$ as the stack of all actuated joint positions,  $\boldsymbol{\theta}_{l}$ for all four lateral joint positions, $\boldsymbol{\theta}_{t}$ for all eight transversal joint positions (two per leg), and $\boldsymbol{\theta}^*_{\triangle},~\triangle\rightarrow m,l,t,$ denoting the desired quantities. The ankle joint placement on the inner and outer shank is denoted $P_{ia}$ and $P_{oa}$, respectively. For kinematic analysis, we virtually separate the ankle joint, though mechanically  $P_{ia} = P_{oa}$

\begin{table}[t]
    \caption{Key Robot Parameters}
    \vspace{-2ex}
    \label{tab:robot_params}
    \centering
    \begin{tabular}{lclc} \thickhline
        \textbf{Parameter} & \textbf{Value} & \textbf{Parameter}  & \textbf{Value}\\ \hline
        Robot weight & \SI{14.5}{\kilogram} &  Nominal height & \SI{0.35}{\meter} \\
        Body length & \SI{0.67}{\meter} & Lateral peak torque & \SI{18.0}{\newton\meter} \\
        Body width & \SI{0.38}{\meter} &   Transversal peak torque & \SI{24.8}{\newton\meter} \\
        Thigh length  & \SI{0.175}{\meter} & Lateral $\boldsymbol{\theta}_m^{def}$ & \SI{45}{\degree}  \\
        Shank length & \SI{0.3}{\meter} & Transversal $\boldsymbol{\theta}_m^{def}$& \SI{0}{\degree}\\ 
         \thickhline
    \end{tabular}
    \vspace{-1.5ex}
\end{table}

\begin{figure}[t]
    \centering
    \includegraphics[width=0.485\textwidth]{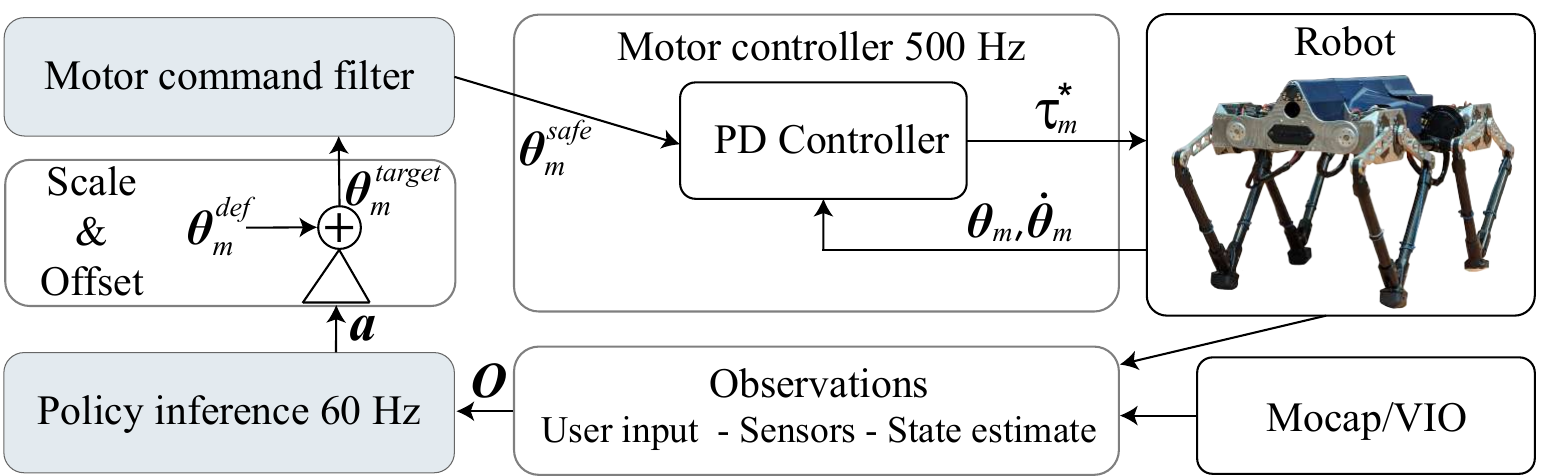}
    \vspace{-4ex}
    \caption{Control setup in simulation and on robot.}
    \vspace{-3ex}
    \label{fig:control_loop}
\end{figure}

The system utilizes the NVIDIA Jetson Orin NX as its onboard computer, which interfaces with the CAN-Bus communication and executes the PD motor controllers at \SI{500}{\hertz}, while the RL policy inference runs at \SI{60}{\hertz}. Fig. \ref{fig:control_loop} illustrates the control setup used to deploy the different policies on the robot, with observations $\mathbf{o}$ and policy actions $\mathbf{a} \in [-1,1]$. Actions are rescaled and interpreted as offsets to nominal motor positions $\boldsymbol{\theta}_m^{def}$, with task-specific rescaling. The target motor angles are computed as $\boldsymbol{\theta}_m^{target} = \mathbf{a} \odot \mathbf{s} +\boldsymbol{\theta}_m^{def}$, where $\mathbf{s}$ represents the action scales, and $\odot$ denotes element-wise multiplication. A motor command filter then processes the target motor angles to produce safe motor targets $\boldsymbol{\theta}_m^{safe}$, which are sent to the PD controllers to generate commanded torques $\boldsymbol{\tau}^*_m$. For walking, all motors use \SI{60}{\degree} as action scales to allow for an appropriate range of motion, while for jumping, lateral motors use \SI{15}{\degree}, and transversal motors use \SI{90}{\degree} to enable larger errors in the motor position targets and generate higher commanded torque from the PD controllers.

\label{sec:motor_command_filter}
We now detail the motor command filter shown in Fig. \ref{fig:control_loop}. The motor position targets are filtered before being sent to the PD controller to ensure safe motion with minimal risk of collision. The actuated joint references are constrained to specific joint limits, $ \boldsymbol{\theta}_{min} \leq \boldsymbol{\theta}_m \leq \boldsymbol{\theta}_{max}$. Additionally, due to the five-bar linkage leg design, we have the additional constraint $ l \leq \theta_{it} + \theta_{ot} \leq u$, where $l$ and $u$ are the lower and upper bounds for the sum of transversal joint angles. The simplest filter would be to always constrain the motor setpoint within these ranges, however, doing so would necessitate very aggressive PD gains in order to obtain the torque bandwidth required to execute dynamic maneuvers, such as jumping. Therefore, inspired by \cite{atanassov2024curriculumbased}, we implement a predictive filtering approach that calculates the time-to-violation based on current joint positions and velocities. The filter then gradually enforces the limits as they are approached. This enables the policy to exhibit close to maximum torque across the whole operating range, which is crucial for jumping, without sacrificing position control's safety and learning capabilities.

\section{KINEMATIC CONSTRAINTS}
\label{sec:sim_kin}

\label{sec:kin}
Due to \robotname{}'s five-bar linkage leg design, not all configurations are valid kinematic states. By virtually cutting open the closed kinematic chain (CKC), following \cite{balajidynamics}, we can formulate the following kinematic constraint,

\small
 \begin{equation}
    \label{eq:ckc}
    \mathbf{C}_{ckc}\left(\mathbf{q}\right) \doteq \ 
     ^{\mathcal{M}}\mathbf{p}_{ia}\left(\mathbf{q}\right) - \   ^{\mathcal{M}}\mathbf{p}_{oa}\left(\mathbf{q}\right) = \mathbf{0},
 \end{equation}
\normalsize

\noindent where $\mathbf{q}$ denotes the complete robot configuration vector including base pose and all joint positions, $\mathcal{M}$ denotes the motor housing frame, while $^{\mathcal{M}}\mathbf{p}_{ia}$ and $ ^{\mathcal{M}}\mathbf{p}_{oa}$ denote the vector from an arbitrary reference point to the inner and outer ankle joint.  One such constraint is formulated for each leg module and stacked together. Note that by design (see  Fig. \ref{fig:leg_config}), Equation \ref{eq:ckc} is always fulfilled in the $y$ coordinate, making the problem planar. 

We choose to treat Equation \ref{eq:ckc} numerically, necessitating the computation of the first-order derivatives. Computing kinematic derivatives of an articulated system like \robotname{} is a well-studied problem. To express these quantities in the motor housing frame, \textit{Euler's rule of differentiation} is used

\vspace{-1.75ex}

\small
\begin{equation}
    \label{eq:jac_ckc}
    \mathbf{J}_{ckc} \doteq \frac{ ^{\mathcal{M}}\partial\mathbf{C}_{ckc}}{\partial \mathbf{q}} =\  ^{\mathcal{M}} \!  \mathbf{J}_{p_{ia}} - \ ^{\mathcal{M}} \! \mathbf{J}_{p_{oa}} + \left[\mathbf{C}_{ckc}\right]_{\times} \!^{\mathcal{M}}\mathbf{J}_{rot},
\end{equation}
\normalsize

\noindent where  $^\mathcal{M}\mathbf{J}_{p_{ia}}$ and $^\mathcal{M}\mathbf{J}_{p_{oa}}$ correspond to the translational part of geometric Jacobians of the inner and outer ankle joint respectively, $[\cdot]_{\times}$ denotes the skew-symmetric matrix operator, and  $^\mathcal{M}\mathbf{J}_{rot}$ is the rotational part of the geometric Jacobian of the motor housing. Additionally, the last term in Equation \ref{eq:jac_ckc} is equivalent to zeroing out all columns of the Jacobians not corresponding to the leg joint indices.

Stacking the CKC constraint  from Equation \ref{eq:ckc} for each leg and, in addition, considering the current and desired paw positions, $^\mathcal{B}\mathbf{p}_{paw}$ and $^\mathcal{B}\mathbf{p^*}_{\!\!\!paw}$, yields the following total kinematic constraint

\vspace{-0.75ex}

\small 
\begin{equation}
    \label{eq:kinemtic_consraint}
    \mathbf{C}\left(\mathbf{q}\right) \doteq \left[\begin{array}{cc}
         &^\mathcal{M}\mathbf{C}_{ckc}  \\ &^\mathcal{B}\mathbf{p}_{paw} \  - \ ^{\mathcal{B}}\mathbf{p^*}_{\!\!\!paw}\\ 
    \end{array} 
    \right],
\end{equation}
\normalsize

\noindent with $\mathcal{B}$ denoting the body frame. In addition, we can obtain the full constraint Jacobian $\mathbf{J}_c$   by stacking the Jacobian of each term. This enables applying a classical weighted inverse kinematic (IK) scheme. Using JAXSIM \cite{ferigo_jaxsim_2022} enables parallel deployment across thousands of environments on the GPU.

\section{LEARNING FRAMEWORK}
\label{sec:DRL_pipeline}
We train walking and jumping policies for \robotname{} using separate reward functions, each comprising a sum of terms. All policies are also subject to a series of regularization rewards, which encourage smoother and safer motions. These rewards are listed in Table \ref{tab:rew_reg}. For brevity, we introduce the following notation for the Exponential and Laplacian Kernel, respectively: $\phi_\sigma(x) := \exp(-\frac{x^2}{\sigma^2})$ and $\psi_\sigma(x) := \exp(-\frac{|x|}{\sigma})$.

\subsection{Walking}
\label{sec:walking}
Although the focus of this work is on jumping policies, walking is also facilitated for completeness. In this case, the objective of \robotname{}'s RL-based locomotion policy is to achieve accurate velocity tracking performance. The policy is rewarded by tracking the command vector, consisting of commanded linear velocities in $x$, $y$, and yaw rate: $^\mathcal{B}\boldsymbol{c} =  [v_x^{*}, v_y^ {*},  \omega_z^{*}]$. The observations for the walking policy are 

\vspace{-2.75ex}

\begin{eqnarray}
\mathbf{o} = [^\mathcal{B}\mathbf{v},~^\mathcal{B}\boldsymbol{\omega},~^\mathcal{B}\mathbf{g},~ ^\mathcal{B}\mathbf{c},~\boldsymbol{\theta}_m^{rel},~\dot{\boldsymbol{\theta}}_m,~\mathbf{a}^{t-1}], 
\end{eqnarray}
where $^\mathcal{B}\mathbf{v}$ is the body linear velocity,
$^\mathcal{B}\boldsymbol{\omega}$ is the body angular velocity,
$^\mathcal{B}\mathbf{g}$ is the gravity vector,
$\boldsymbol{\theta}_m^{rel}$ are the joint positions relative to default,
$\dot{\boldsymbol{\theta}}_m$ are the joint velocities, and 
$\mathbf{a}^{t-1}$ are the previous actions.

Table \ref{tab:rew_walking}  summarizes the rewards used during training. The \textit{Linear velocity tracking} and \textit{Yaw rate tracking} rewards track the desired velocity command while the \textit{Vertical velocity L2} reward avoids movement along the undesired axis. To achieve a more natural gait, the \textit{Stand}, \textit{Lateral position}, and \textit{Transversal position} rewards regularize joint positions around nominal values close to the default positions. Different targets and reward shapes for moving versus stationary states are used to distinguish between moving and standing. In addition to the rewards listed in Table \ref{tab:rew_walking}, we incorporate two rewards that encourage longer steps and lower gait frequency, inspired by \cite{rudin2022learning}. Additionally, any collisions of the robot with itself or the environment are penalized, while any collision involving the robot's base causes termination. To increase the robustness of the policy, we randomize the initial configuration of the robot by sampling random base poses and paw positions. Using the IK scheme presented in Section \ref{sec:kin}, the corresponding robot state is calculated.

\begin{table}
\caption{Regularization Rewards}
    \vspace{-2ex}
    \label{tab:rew_reg}
    \centering
    \begin{tabular}{ll} \thickhline
        \multicolumn{2}{c}{Reward formulation} \\
        \hline
         Action clip & $||\boldsymbol{\theta}_m^{target} - \boldsymbol{\theta}_m^{safe}||^2$  \\ 
         Motor torque & $ ||\boldsymbol{\tau}_m||^2 $ \\ 
         Joint acceleration & $||\boldsymbol{\Ddot{\theta}}_m||^2 $  \\ 
         Action rate & $ ||\boldsymbol{a}^{t} - \boldsymbol{a}^{t-1}||^2$  \\ 
         Jerk & $||\mathbb{I}[\text{sgn}(\boldsymbol{\tau}^{t}_m) \neq \text{sgn}(\boldsymbol{\tau}_m^{t-1})]||_1$  \\ \thickhline
    \end{tabular}
    \vspace{-2ex}
\end{table}

\begin{table}
    \caption{Walking Rewards}
    \vspace{-2ex}
    \label{tab:rew_walking}
    \centering
    \begin{tabular}{ll} \thickhline
        \multicolumn{2}{c}{Reward formulation} \\
        \hline
         Linear velocity tracking   &  $\phi_{\sigma_1}(||^\mathcal{B}\mathbf{v}_{xy}-\ \! ^\mathcal{B}\mathbf{c}_{xy}||)$  \\ 
         Yaw rate tracking &   $\phi_{\sigma_2}(^\mathcal{B} \omega_{z}-\ ^\mathcal{B} \omega_z^*)$  \\  
         Vertical velocity L2 & $ v_z^2$ \\  
         Lateral stability L2 &   $||^\mathcal{B}\boldsymbol{\omega}_{xy}||^2$ \\  
         Flat L2 & $ ||^\mathcal{B}\mathbf{g}_{xy}||^2$  \\  
         Stand &  $\phi_{\sigma_3}(||\boldsymbol{\theta}_{t} - \boldsymbol{\theta}_{t}^{*}||)$  \\  
           Lateral position& 
         $ \phi_{\sigma_4}(||\boldsymbol{\theta}_{t} - \boldsymbol{\theta}_{t}^{*}||^{4}) -1 $ \\  
         Transversal position & 
         $ \phi_{\sigma_5}(||\boldsymbol{\theta}_{l} - \boldsymbol{\theta}_{l}^{*}||^{10}) -1 $
         \\ \thickhline
    \end{tabular}
    \vspace{-5ex}
\end{table}

\subsection{Vertical Jumping}
\label{sec:vertical-jump}
The vertical jumping policy is trained to execute controlled jumps that reach a specified target height and achieve a safe landing. The observation vector is given by 

\vspace{-3ex}
\begin{eqnarray}
\mathbf{o} = [h^*,~c, ~ h,  ~  ^\mathcal{B}\mathbf{v},~^\mathcal{B}\boldsymbol{\omega},~^\mathcal{B}\mathbf{g},~\boldsymbol{\theta}_m^{rel},~\dot{\boldsymbol{\theta}}_m,~\mathbf{a}^{t-1}],
\end{eqnarray}
where $h^*$ is the desired jump height,
$c\in \{0, 1\}$ is the jump command,
$h$ is the current base height,
and the remaining terms are the same as in Section \ref{sec:walking}. The jumping command $c$ is necessary to synthesize a difference in the robot's state before and after a vertical jump is performed.

The rewards for the RL training are listed in Table \ref{tab:rew_jump}. Note that not all rewards are necessarily applicable at all stages of a jump. We therefore track the agent's \textit{jump state} (\textit{stance, in-flight, landed}), and apply the different rewards where applicable. The main reward is the \textit{Jump height} reward, which rewards the agent based on the error between the maximum achieved jump height $h_{max}$, and the desired jump height $h^*$. This reward is handed out discontinuously, once per jump. To achieve a denser reward signal, we also estimate the anticipated jump height $\hat{h}_{max}$ continuously between takeoff and when the peak jump height is achieved using the projectile equations of motion for a single rigid body. We refer to this term as the \textit{Est jump height} reward. In addition, motion dynamics indicate that vertical jumping maneuvers should exhibit strong symmetry, which we encourage through the \textit{Joint Symmetry} reward. The policy is also encouraged to minimize the angular velocity of the base while keeping a flat orientation, where $\boldsymbol{\alpha}_{error}$ represents the rotation error relative to the desired orientation. Furthermore, the agent is biased towards predefined desired joint positions while airborne and after landing. Note that before and during takeoff, no such reference signals are provided. On real hardware, as opposed to simulation, there is a chance of mechanical strain and failure when executing jumping maneuvers. To address this, several rewards are introduced to minimize mechanical stress. Ground contact forces $\mathbf{F}_{ground}$ are regularized to reduce actuator loading, while \textit{Soft impacts} are encouraged by rewarding deceleration in the direction of motion, using body acceleration $\mathbf{a}_{body}$, L2 normalized body velocity $\tilde{\mathbf{v}}_{body}$, and a maximum acceleration threshold $a_{max}$. Finally,  the first \SI{0.3}{\second}  after touchdown, the robot is encouraged to retract its legs (\textit{Damp landing}) while letting the base fall downwards (\textit{Catch landing}).  

\begin{table}
    \caption{Jumping Rewards}
    \vspace{-2ex}
    \label{tab:rew_jump}
    \centering
    \begin{tabular}{llc} \thickhline
        \multicolumn{2}{c}{Vertical jump rewards} \\
        \hline
        Jump height & $\phi_{\sigma_6}(h_{max} - h^*) + 3 \cdot \psi_{\sigma_7}(h_{max} - h^*))$\\ 
        Est jump height & $\phi_{\sigma_8}(\hat{h}_{max} - h^*) + 3 \cdot \psi_{\sigma_9}(\hat{h}_{max} - h^*))$\\ 
        Joint symmetry  & $ \phi_{\sigma_{10}}(\text{Var}(\boldsymbol{\theta}_{t})) \cdot \phi_{\sigma_{11}}(||\boldsymbol{\theta}_{l}||)$ \\ 
        \thickhline
        \multicolumn{2}{c}{Horizontal jump rewards} \\ \thickhline
        Tracking&   $\phi_{\sigma_{12}}\left(\mathbf{|e|}\right)$\\ 
        Est tracking&  $\phi_{\sigma_{13}}\left(\mathbf{|\hat{e}|}\right) + 0.1\cdot\phi_{\sigma_{14}}\left(\mathbf{\hat{e}}\right))  $ \\
        Joint symmetry  & $ \phi_{\sigma_{15}}(||\boldsymbol{\theta}_{\text{left transversal}}-\boldsymbol{\theta}_\text{right transversal}||)$
        \\  \thickhline
        \multicolumn{2}{c}{Common rewards} \\
        \thickhline
        Angular velocity &  $ \phi_{\sigma_{16}}(||^\mathcal{B}\boldsymbol{\omega}||)$\\
         Orientation error &  
        $ \phi_{\sigma_{17}}(\boldsymbol{\alpha}_{error}^2)$\\ 
         Desired joint pos  &  $
        \phi_{\sigma_{18}}(||\boldsymbol{\theta}_m - \boldsymbol{\theta}_m^{*}||)$\\ 
        Ground force L2   &  $||\mathbf{F}_{ground}||^2$ \\
        Soft impact &   $\max(0, 1 - |\min(0, \frac{\mathbf{a}_{body}}{a_{max}} \cdot \tilde{\mathbf{v}}_{body})|)$ \\ 
        Catch landing &  $ \text{clamp}(-v_z, 0, 1) $ \\
        Damp landing &   $\text{clamp}(\dot{\boldsymbol{\theta}}_{t}, 0, 1) $\\ \thickhline
    \end{tabular}
    \vspace{-4ex}
\end{table}

\subsection{Horizontal Jumping}

The horizontal jumping policy is tasked to steer the center of the quadruped to a position in the plane $\mathbf{p}^*$ by performing a jump. The observation vector is given by 

\vspace{-3ex}
\begin{eqnarray}
\mathbf{o} = [\mathbf{R}^T_{yaw}\mathbf{e},~h,~^\mathcal{B}\mathbf{v},~^\mathcal{B}\boldsymbol{\omega},~^\mathcal{B}\mathbf{g},~\boldsymbol{\theta}_m^{rel},~\dot{\boldsymbol{\theta}}_m,~\mathbf{a}^{t-1}],
\end{eqnarray}
where $\mathbf{R}_{yaw}$ is the 2D rotation matrix corresponding to the yaw component of the base orientation, and $\mathbf{e}$ is the horizontal (x and y) component of the tracking error $\mathbf{p}^*-\mathbf{p}$ between target and current robot position, $\mathbf{p}^*$ and $\mathbf{p}$, respectively. The multiplication with the transpose of the rotation matrix effectively expresses the tracking error in the robot's yaw-aligned frame. The remaining terms have already been described in Sections \ref{sec:vertical-jump} and \ref{sec:walking}.

The horizontal jumping rewards involve mostly adaptations to those used for vertical jumping and are listed in Table \ref{tab:rew_jump}. The main difference is the \textit{Tracking} reward, which acts as a navigation reward driving the robot to the desired position. This reward is applied continuously, but the signal strength naturally decreases as the distance to the goal increases. To address this, we make the reward signal denser by estimating the final landing position using the projectile equations of motion while the robot is in-flight and applying the \textit{Est tracking} reward based on the estimate of the error at the end of the jump $\mathbf{\hat{e}}$.

\subsection{Curriculum-based Reference State Initialization}

\begin{figure}[t]
     \centering
     \includegraphics[width=0.485\textwidth]{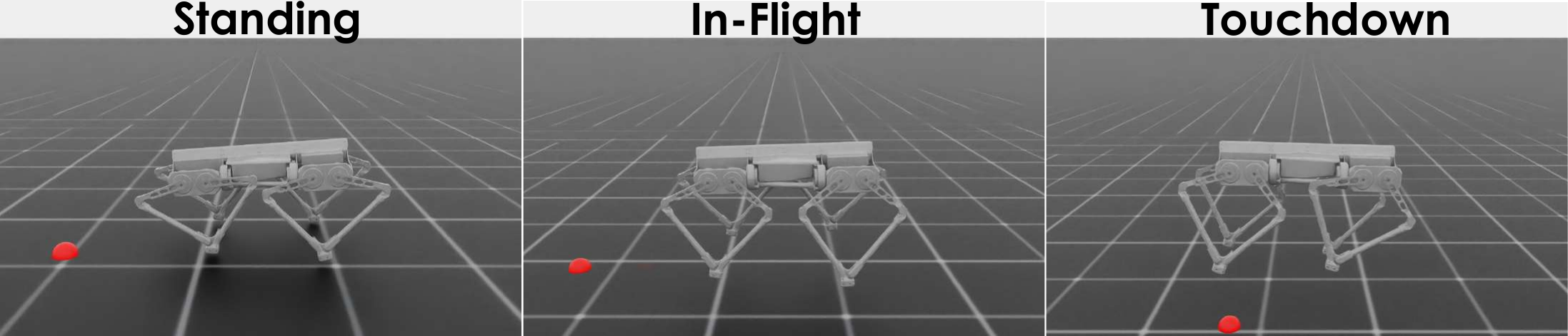}
     \vspace{-4ex}
     \caption{Examples of RSI for different stages of a jump, \textit{standing}, \textit{in-flight}, and \textit{touchdown}. The red dot indicates the desired landing position.}
     \label{fig:init_sim}
     \vspace{-4ex}
\end{figure}

To help the agent learn how to jump, an elaborate reference state initialization (RSI) scheme is employed, where the agents are initialized randomly at the different stages of a jumping maneuver, namely \textit{standing/squatting}, \textit{in-flight}, just before \textit{touchdown}, and  \textit{landed} close to the goal position, as illustrated in Fig. \ref{fig:init_sim}. To generate the base state of the airborne agents we sample base positions and velocities along projectile trajectories corresponding to the desired jump. Section \ref{sec:kin}, joint configurations are simultaneously randomized. Especially for horizontal jumping, the agents that are initialized in \textit{touchdown} have their paws positioned slightly forward, which helps the policy learn an effective bracing strategy. Robots initialized in \textit{standing} begin with randomized standing height and base orientation, initially favoring deep squatting positions in early training, then expanding to cover broader standing configurations as performance improves. For horizontal jumps, the robots are also initialized with a slight forward pitch. As the training proceeds, randomization occurs over all reasonable standing configurations. This transition is controlled by performance-based curriculum thresholds. A separate curriculum progressively increases the commanded jump distance/height. The RSI scheme removes the need for auxiliary rewards such as squat depth used in \cite{atanassov2024curriculumbased}, where, instead of forcing a reference behavior, we help the agent explore efficient strategies.

To further improve the training pipeline, we introduce multiple termination criteria to prevent unwanted behavior and guide the agent away from unproductive states. These include terminating the agent if it has not performed a jump after a specified time limit, or if the predicted or measured jumping performance falls below defined thresholds. The agent is also terminated by any collisions with itself or the environment, or if the deceleration during touchdown is above a set threshold. To force jumping behavior, the agent is terminated if the base has moved beyond a threshold distance from the starting position without performing a jump.

\subsection{Neural Architectures and Implementation}
We represent the policies using a three-layer Multilayer Perceptron (MLP) with  Exponential Linear Unit (ELU) activations. The layer widths are $[512,256,128]$  for walking and $[256,128,128]$ for jumping. IsaacLab~\cite{mittal2023orbit} provides the simulation environment, while RL Games~\cite{rl-games2021} implements the PPO algorithm. Training is parallelized across $4096$ environments on an NVIDIA RTX 3090.

\subsection{Crossing the Sim2Real Gap}

Deploying the learned policy on real hardware requires bridging the Sim2Real gap. In this work, the three main strategies employed to achieve this are: a) \textit{System identification:} improving simulation accuracy to reflect the actual dynamical behavior of the quadruped through motor parameter identification and actuator characterization; b) \textit{Domain randomization:} incorporating noise and parameter uncertainty during training for walking and jumping policies; c) \textit{Comprehensive state coverage:} All policies are initialized across all reasonable configurations, enabling comprehensive state coverage of the system's operational envelope during training. Tables \ref{tab:DR} and \ref{tab:noise} detail the domain randomization variables and the variance of the Gaussian noise applied during training. For jumping, the latency randomization is only applied to actions, but for walking, it is also applied to observations.

\begin{table}
    \caption{Randomization Variables}
    \vspace{-2ex}
    \label{tab:DR}
    \centering
    \begin{tabular}{lcc} \thickhline
         \textbf{Variable} & \textbf{Walking} & \textbf{Jumping} \\ \hline
         Static friction            & [0.8, 0.95] & [0.9, 1.2] \\ 
         Dynamic friction           & [0.7, 0.8] & [0.8, 0.9] \\ 
         Base mass [\si{\kilogram}]            & [-1.0, 2.0] & [-1.0, 2.0] \\ 
         Link mass           & $\pm20\%$ & $\pm20\%$ \\ 
         Center of Mass shift [\si{\meter}]             & [-0.03, 0.03] & [-0.03, 0.03] \\
         Actuator gains      & $\pm40\%$ & $\pm40\%$ \\
         No-load speed scaled       & [0.6, 1.2] & [0.8, 1.2] \\ 
         Cutoff speed  scaled       & [0.6, 1.4] & [0.8, 1.4] \\ 
         Motor friction [\si{\newton\meter\second}] & [0.0, 0.04] & [0.005, 0.04]\\
         Motor armature     & $\pm40\%$ & $\pm40\%$ \\ 
         Joint offsets [deg]         & [-2.0, 2.0] & [-2.0, 2.0] \\
         Latency    [\si{\milli\second}]              & [0.0, 32.0] & [0.0, 16.0] \\
         External force [\si{\newton}]                           & [-10.0,10.0] & [-5.0, 5.0] \\
         External torque [\si{\newton\meter}]                        & [-3.0, 3.0] & [-3.0, 3.0] \\
         \thickhline
    \end{tabular}
    \vspace{-2ex}
\end{table}

\begin{table}[t]
    \caption{Observation Noise}
    \vspace{-2.5ex}
    \label{tab:noise}
    \centering
    \begin{tabular}{lcc} \thickhline
         \textbf{Variable} & \textbf{Walking} & \textbf{Jumping} \\ \hline
         Body linear velocity [\si{\meter\per\second}] & 0.1 & 0.1 \\
         Body angular velocity [\si{\per\second}] & $0.1$ & $0.1$  \\ 
         Projected gravity [\si{\meter\per\second\squared}] & 0.5 & 0.5 \\
         Joint position [deg]      & $3$ & $3$ \\ 
         Joint velocity [deg]      & $10$ & $10$\\ 
         \thickhline
    \end{tabular}
    \vspace{-3ex}
\end{table}

\section{SIMULATION STUDIES}
\label{sec:sim_studies}
We conduct a series of simulation studies to assess the trained policies across wide variations of states and inputs. For all tests, a successful jump execution is defined as reaching within \SI{0.1}{\meter} of the commanded target height/position. All jumping tests in simulation are performed with a torque saturation of \SI{18}{\newton\meter}.

\subsection{Vertical jump}

\begin{figure}[t]
    \centering
    \vspace{-1ex}
    \includegraphics[width=0.49\textwidth]{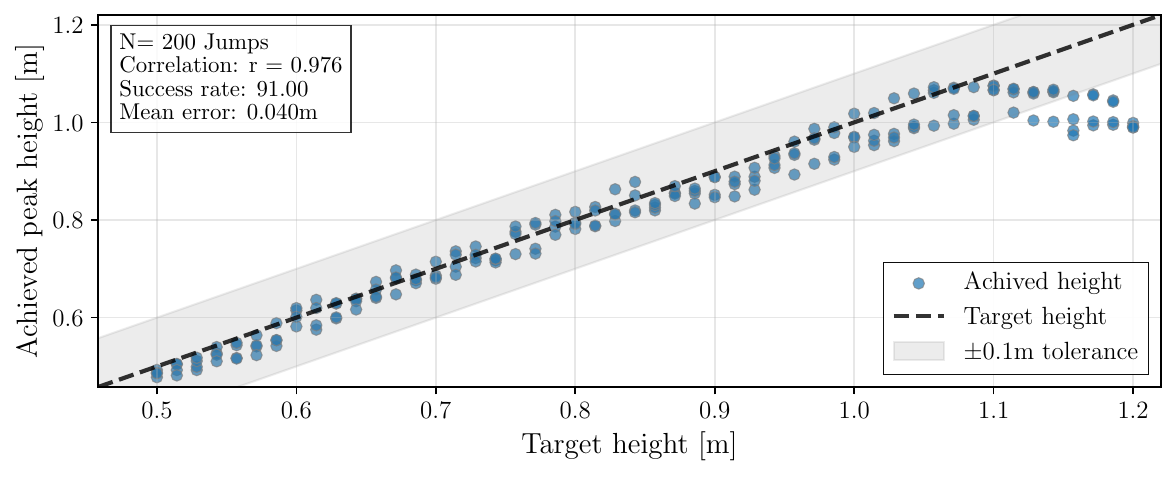}    
    \vspace{-5ex}
    \caption{Achieved vertical jump height vs target height in simulation. Training range: \SIrange{0.6}{1.1}{\meter}.}
    \label{fig:vertical_corrolation_plot}
    \vspace{-4ex}
\end{figure}

To validate the capabilities of the trained vertical jumping policy, we simulate \num{200} jumps with target heights spanning from \SI{0.5}{\meter} to \SI{1.2}{\meter}, which exceeds the commands observed during training by $\pm$\SI{0.1}{\meter}. Fig. \ref{fig:vertical_corrolation_plot} shows the high tracking performance of the policy with a mean absolute tracking error of \SI{0.04}{\meter}. The policy demonstrated generalization slightly beyond the training range, achieving $100$\% success rate for heights between \SIrange{0.5}{1.1}{\meter}. Performance declined only at the upper extreme where the 
torque limits are reached, resulting in an overall \num{91}\% success rate across the full \SIrange{0.5}{1.2}{\meter} test range.

\subsection{Horizontal Jump}

\begin{figure}[t]
    \centering
    \includegraphics[width=0.49\textwidth]{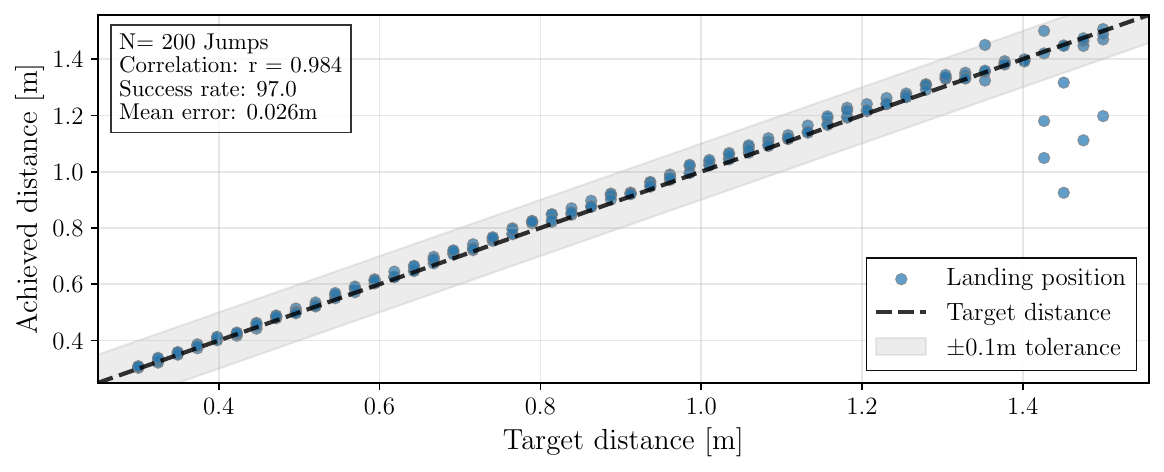}    
    \vspace{-5.5ex}
    \caption{Achieved forward jump distance vs target distance (no lateral/y component).  Training range: \SIrange{0.4}{1.0}{\meter}.}
    \label{fig:straight_forward_corrolation_plot}
    \vspace{-2ex}
\end{figure}

We evaluated the horizontal jumping policy by simulating \num{200} jumps with target landing position covering forward distances of \SIrange{0.3}{1.5}{\meter}, which is outside the training distribution by $-$0.1 and $+$\SI{0.5}{\meter}, respectively. Fig. \ref{fig:straight_forward_corrolation_plot} shows high tracking performance with a mean tracking error of \SI{0.026}{\meter} and \num{97}\% overall success rate, achieving a \num{100}\% success rate up to \SI{1.4}{\meter}, thus showing strong generalization. Tracking performance decreases as the target approaches the extreme distances at the system limits. 

\begin{figure}[t]
    \centering
    \includegraphics[width=0.49\textwidth]{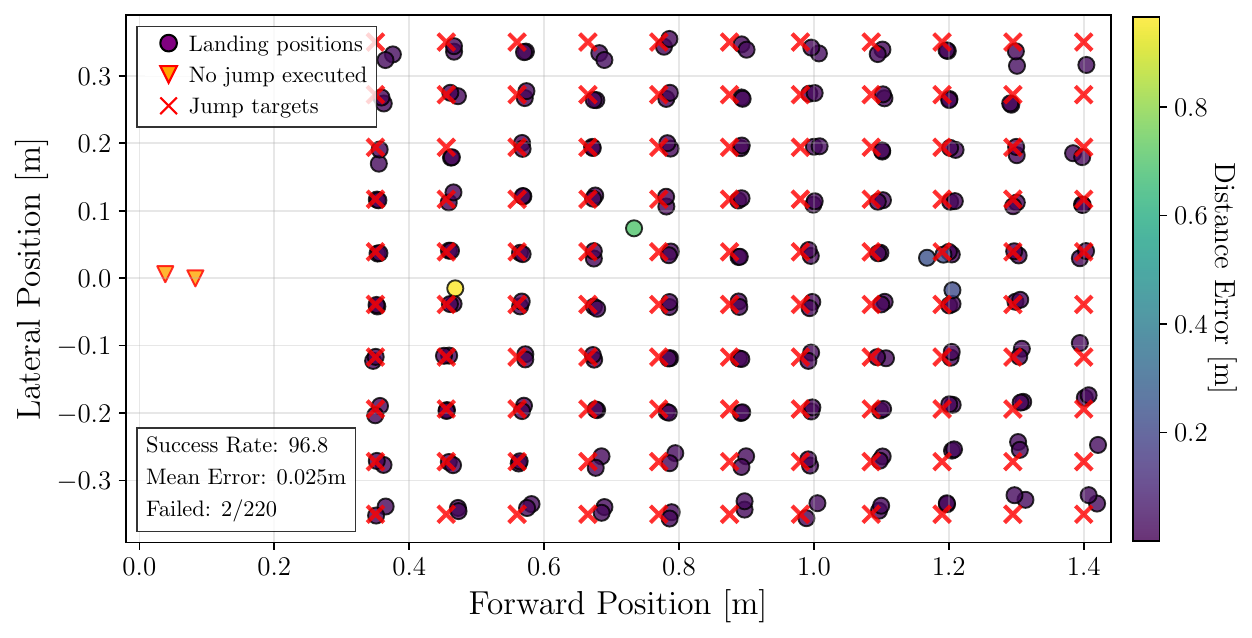}    
    \vspace{-5.5ex}
    \caption{Jump landing positions vs jump targets in simulation. The ${\color{red}\boldsymbol{\times}}$ showcases the grid of jump targets. The color of $\circ$ represents the distance from the target. Training range is x $\in$ [\num{0.4}, \num{1.0}]~\si{m} and y $\in$ [$-$\num{0.3}, \num{0.3}]~\si{m}.}
    \label{fig:forward_targetplot}
    \vspace{-4ex}
\end{figure}

Diagonal jumping precision was tested using a grid of target positions with forward distances ranging from  \SIrange{0.35}{1.4}{\meter}, each combined with a sideways component varying from $-$\SIrange{0.35}{0.35}{\meter}. Two jumps were executed per target location. Fig. \ref{fig:forward_targetplot} shows that the strong tracking performance persists when considering diagonal jumping, achieving a mean tracking error of \SI{0.025}{\meter}  and a \num{96.8}\% success rate. All failures occurred at the \SI{1.4}{\meter} distance, consisting of two cases where no jump was executed and five unsuccessful jumps that fell short of the target, with all failures attributed to the distance far exceeding what was seen during training.

\section{EXPERIMENTAL VALIDATION}
\label{sec:experimental_validation}
This section covers the experimental validation and testing of the trained DRL policies. For the walking policy, tests are conducted both with and without onboard state estimation, while for the jumping policies, a motion capture (Mocap) system is used. All walking and jumping tests were conducted with a torque saturation of \SI{18}{\newton\meter} to ensure safer operation.

\subsection{Walking}
\begin{figure}[t]
    \centering
    \includegraphics[width=0.47\textwidth]{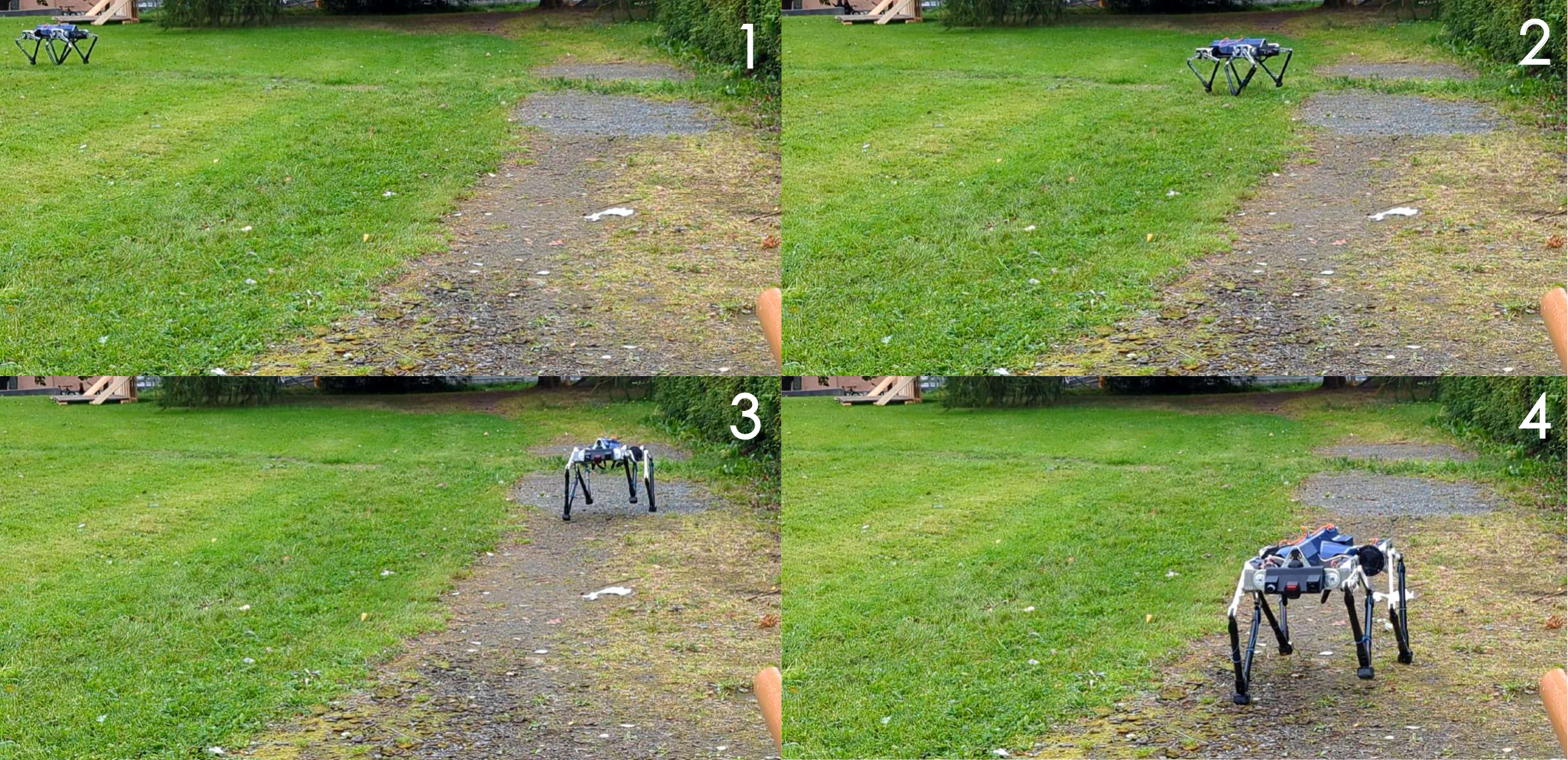}    
    \vspace{-2ex}
    \caption{\robotname{} walking outside on 1) grass, 2) uneven grass, 3) gravel path, 4)  dirt path.}
    \label{fig:walking_comp}
    \vspace{-1.5ex}
\end{figure}

\begin{figure}[t]
    \centering
    \includegraphics[width=1.00\linewidth, trim={0.02\linewidth} {0.02\linewidth} {0.02\linewidth} {0.02\linewidth}, clip]{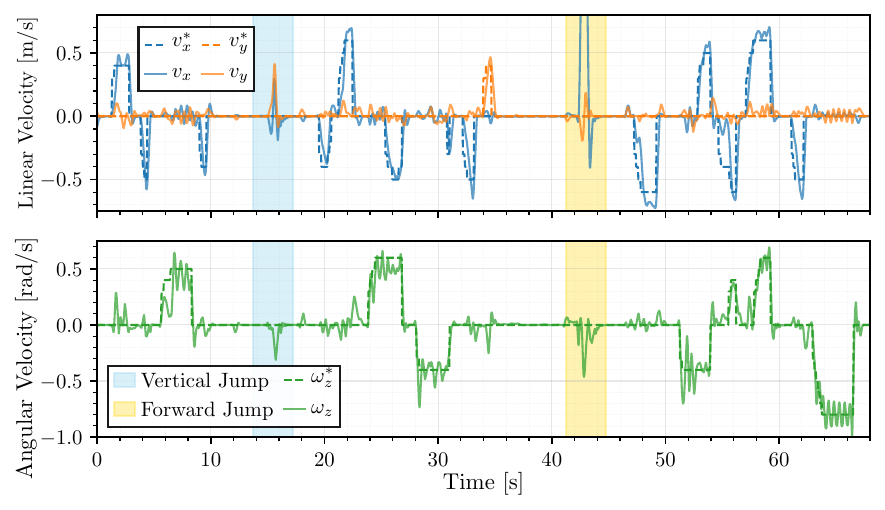}
    \vspace{-5ex}
    \caption{Walking policy commanded vs actual velocity, tracking performance during combined test with vertical and forward jump maneuvers.}
    \vspace{-3ex}
    \label{fig:walk_jump_walk}
\end{figure}

We validate the walking policy $\pi_{W}$ across diverse terrain conditions using two state estimation approaches. Indoor tests utilized Mocap for body state feedback, while outdoor validation employed a custom VIO system consisting of a VectorNav VN100 IMU and stereo FLIR Blackfly S GigE cameras. ROVIO \cite{bloesch2017iterated} provided body state estimates, with MSF \cite{lynen13robust} delivering fused estimates at the IMU rate of \SI{200}{\hertz}.
The policy was tested across varied surfaces: indoor environments included stone flooring, mats, and uneven wooden surfaces, while outdoor testing encompassed stone and dirt paths, gravel surfaces, and uneven grass terrain, as shown in Fig. \ref{fig:walking_comp}. Command inputs ranged from $[v_x^{*}, v_y^{*}] = [-$\num{0.8}, \num{0.8}] \si{\meter\per\second} and  $\omega_z^{*} = [-$\num{0.8}, \num{0.8}] \si{\radian\per\second} during policy testing. Quantitative tracking performance for the walking policy was measured during integrated locomotion tests combining walking and jumping maneuvers. The policy achieved RMSE values of \SI{0.17}{\meter\per\second}, \SI{0.05}{\meter\per\second}, and \SI{0.12}{\radian\per\second} for $v_x$, $v_y$, and $\omega_z$ tracking, this tracking performance is shown in Fig. \ref{fig:walk_jump_walk}.  Qualitative behavior observed in simulation closely matched hardware performance, demonstrating effective Sim2Real transfer.

\subsection{Vertical Jump}

\begin{figure}[t]
    \centering
    \includegraphics[width=0.49\textwidth]{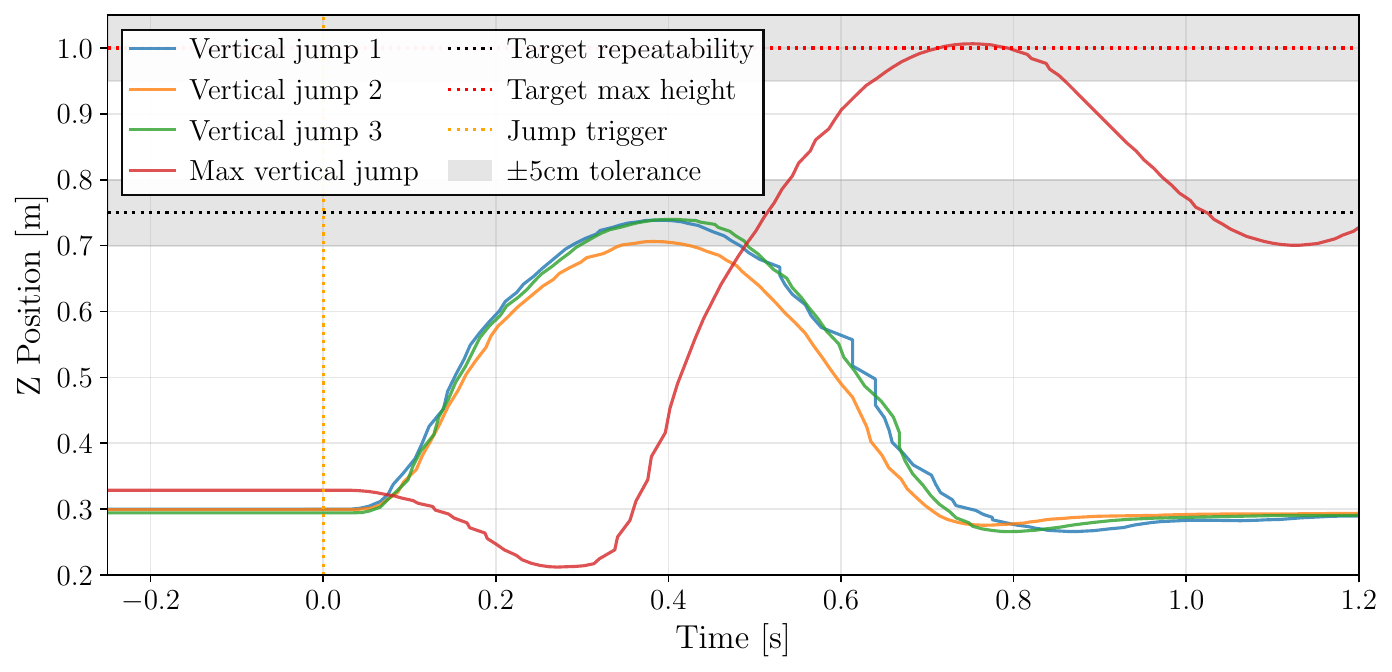}    
    \vspace{-5ex}
    \caption{Jump height vs target height with  \SI{18}{\newton\meter} torque saturation.}
    \label{fig:vertical_comp}
    \vspace{-3ex}
\end{figure}

The vertical jumping policy $\pi_{VJ}$ was first evaluated through three repeated jumps at a target height of \SI{0.75}{\meter}, achieving an average absolute error of \SI{0.023}{\meter} from the target height. The jumping capabilities were also tested at their limits by performing a jump with a target height of \SI{1.0}{\meter}, reaching a height of \SI{1.01}{\meter}. Fig. \ref{fig:vertical_comp} shows the resulting jump trajectories demonstrating jump accuracy. The policy exhibits adaptive behavior, learning to squat deeper as the target jump height increases.  During landing, the policy demonstrated controlled deceleration with minimal base rotation. Fig. \ref{fig:composit_figure} shows the robot performing a vertical jump of \SI{0.75}{\meter}. Note that to prevent excessive motor/gear wear and tear, the robot was caught mid-air with a rope during the maximum height test.

To validate simulation accuracy and establish system performance limits, open-loop experiments were conducted using pre-programmed trajectories that execute maximum squat-jump motions with position commands that saturate motor torque. Using identical control gains and motor filter settings as the RL policy tests, these trajectories achieved a maximum jump height of  \SI{1.075}{\meter} at \SI{18}{\newton\meter} torque saturation. Comparing the maximum open-loop jump of \SI{1.075}{\meter} with the RL policy jump of \SI{1.01}{\meter} demonstrates that the policy achieves \num{94}\% of the theoretical system limits under the set torque constraints.  Simulation studies with stiffer gains, less strict filter settings, and maximum motor torque (\SI{24.8}{\newton\meter}) suggest a maximum possible jump height of \SI{1.42}{\meter} for the system.

\subsection{Horizontal Jump}
We demonstrate the capabilities of the horizontal jump policy $\pi_{HJ}$ through a series of tasks. These include: 1) Max forward jump (\SI{1.25}{\meter}), 2) Forward jump with a lateral component (\SI{0.75}{\meter} forward, $\pm$\SI{0.35}{\meter} sideways), 3) The same jump (\SI{0.85}{\meter} forward, \SI{0.1}{\meter} sideways) repeated three times, 4) Jumping off a \SI{0.15}{\meter} platform, 5) Jumping onto an uneven and movable surface, and 6) Consecutive jumps. 

All experiments were initialized by manually configuring the robot at the desired starting position. Standing in place is then achieved by setting the target position close to the desired starting position. This process naturally introduces some variation in the initial configuration, which is robustly handled by the policy. Jump commands are then generated by specifying target landing positions at various distances relative to the starting location.  Fig. \ref{fig:composit_figure} shows experiments 4, 5, and 6, along with a simulated forward jump of \SI{1.25}{\meter}.

\begin{figure}[t]
    \vspace{1ex}
    \centering
    \includegraphics[width=0.49\textwidth]{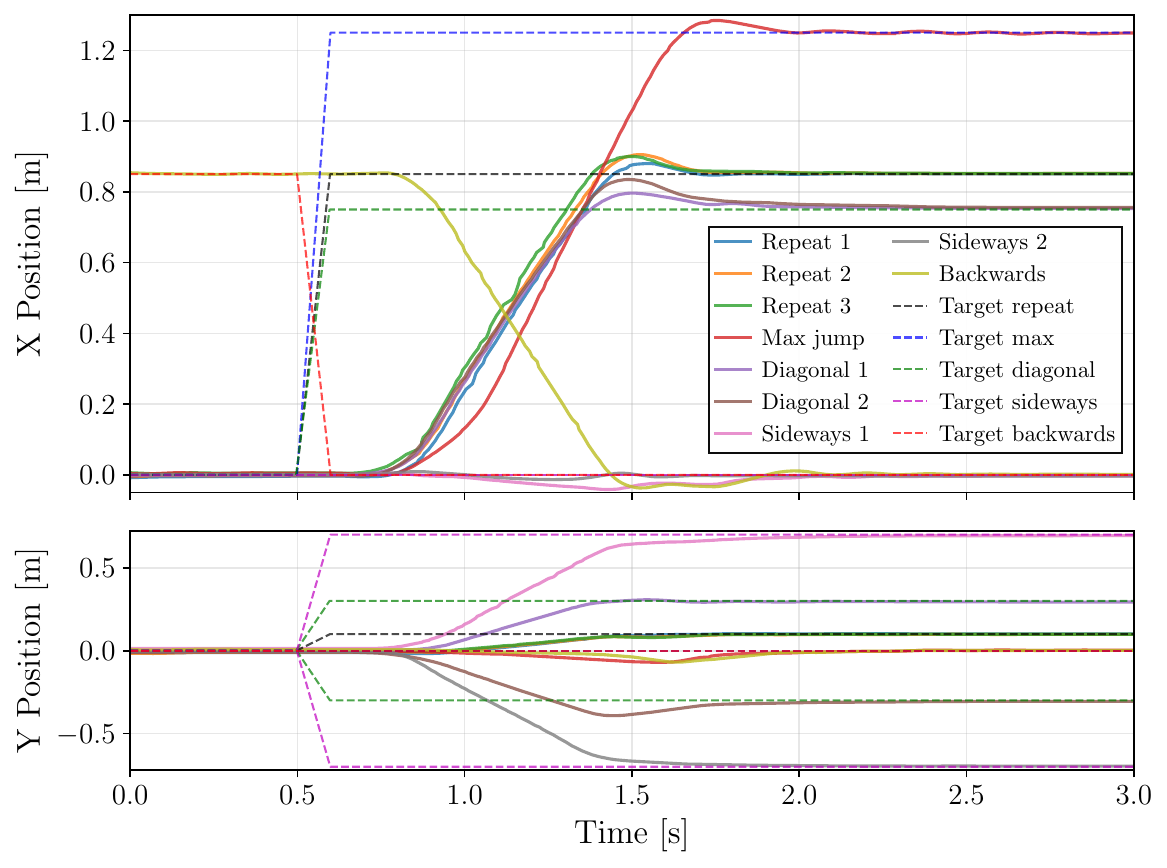}    
    \vspace{-5ex}
    \caption{Jump position vs jump target in x and y.}
    \label{fig:forward_jump}
    \vspace{-4ex}
\end{figure}

The tracking performance for experiments 1, 2, and 3 are presented in Fig. \ref{fig:forward_jump}. The policy achieved a maximum jump distance of \SI{1.25}{\meter} with a landing error of \SI{0.004}{\meter} from the target, which, to the best of the authors' knowledge, exceeds the state-of-the-art for comparable works (see Table \ref{tab:sota} for jump distance comparison). Note that due to variations in hardware, this is not an exact comparison.

\begin{table}[t]
    \caption{Maximum Jump Distance Comparison} 
    \vspace{-2ex}
    \label{tab:sota}
    \setlength{\tabcolsep}{4pt}
    \centering
    \begin{tabular}{lcccccc} \thickhline
         Work: & \cite{caluwaerts2023barkour}& \cite{margolis2021learning}&   \cite{liu2023distance} & \cite{cheng2024extreme} & \cite{atanassov2024curriculumbased} & Ours\\ 
         Jump length [m]: & \SI{0.5}{\meter} & \SI{0.66}{\meter}& \SI{0.8}{\meter} & \SI{0.8}{\meter} & \SI{0.9}{\meter}& \SI{1.25}{\meter}\\ \thickhline
         \end{tabular}
     \vspace{-4ex}
\end{table}

Horizontal jumping experiments demonstrate high precision across multiple target configurations. For the two \SI{0.75}{\meter} forward jumps with lateral targets at $\pm$\SI{0.35}{\meter}, the mean landing error the policy achieved was \SI{0.0015}{\meter}. For the repeatability tests targeting  \SI{0.85}{\meter} forward and a \SI{0.1}{\meter} lateral distance, yields a mean landing error of \SI{0.005}{\meter}. Experiments $4$ and $5$ tested the system's robustness to real-world terrain variations, including elevation changes and surface irregularities. Both scenarios validated the policy's ability to maintain stable jumping performance despite non-ideal ground conditions.

Experiment $6$ evaluated sequential jumping capabilities using two test setups. The diagonal tests consisted of a first jump of \SI{0.8}{\meter} forward with $+$\SI{0.2}{\meter} lateral, followed by a second jump of \SI{0.75}{\meter} forward with $-$\SI{0.2}{\meter} lateral. The straight tests consisted of two consecutive \SI{0.8}{\meter}  forward jumps. The policy achieved a mean landing error of \SI{0.0054}{\meter} across all jumps. Note that the first jump could still be completed successfully if the initial takeoff occurs on low-friction surfaces or with targets near the edge of the policy's capabilities. However, these cases might result in an off-nominal configuration when the robot lands the first jump, which increases the chance of failure for the second jump. This can be mitigated by switching to the walking policy to recover the nominal stance.

Additional experiments demonstrated that the proposed RL pipeline could learn omnidirectional jumping through curriculum modifications alone. Separate policies were trained with sideways and backwards commands included in the curriculum,  successfully achieving these multi-directional jumping movements with high accuracy, as shown in Fig. \ref{fig:forward_jump}. However, this generalization compromised peak jumping performance, indicating a trade-off between specialized performance and omnidirectional versatility.

\subsection{Hierarchical policy deployment}
The integration of multiple policies for complex locomotion was evaluated through a hierarchical test combining walking with vertical (\SI{0.75}{\meter}) and horizontal (\SI{0.85}{\meter}) jumps, using manually triggered transitions between policies. Figure  \ref{fig:walk_jump_walk} shows the walking policy's velocity tracking during this test. This test demonstrates the potential for combining such policies to fully leverage the robot's dynamic locomotion capabilities for complex traversal and navigation tasks.

\section{CONCLUSION}
\label{sec:conclusion}
This work demonstrates the training and deployment of walking and jumping RL policies for the \robotname{} quadruped. Fundamental to training the jumping policies is an elaborate RSI scheme, which is enabled by a GPU-parallelized IK-solver, and densifying the rewards using projectile motion equations. Experimental validation on real hardware shows the capabilities of the trained policies to walk across various terrains with onboard state estimation. The jumping policies exhibit excellent tracking performance, both for vertical and horizontal jumps, with horizontal jumps up to \SI{1.25}{\meter} with centimeter accuracy. The policies successfully handle diagonal jumps, jumping from heights, and onto unstructured and loose terrain. Quantitative analysis, both in the real world and in simulations, demonstrates effective Sim2Real transfer. The jumping policies also demonstrate generalization to jumping maneuvers not seen during training. Additionally, the training pipeline demonstrates the potential for omnidirectional jumping through curriculum modifications.

\addtolength{\textheight}{-12cm}   




\bibliographystyle{IEEEtran}

\bibliography{./2025.bib}

\end{document}